# SVM based Multiclass Classifier for Gait phase Classification using Shank IMU Sensor

**Aswadh Khumar G S[1], Barath Kumar J K[1]**
[1]PSG College of Technology, Department of Robotics and Automation Engg, Coimbatore.

Corresponding author: (e-mail:aswadhkhumar15@gmail.com).

**ABSTRACT** In this study, a gait phase classification method based on SVM multiclass classification is introduced, with a focus on the precise identification of the stance and swing phases, which are further subdivided into seven phases. Data from individual IMU sensors, such as Shank Acceleration X, Y, Z, Shank Gyro X, and Knee Angles, are used as features in this classification model. The suggested technique successfully classifies the various gait phases with a significant accuracy of about 90.3%. Gait phase classification is crucial, especially in the domains of exoskeletons and prosthetics, where accurate identification of gait phases enables seamless integration with assistive equipment, improving mobility, stability, and energy economy. This study extends the study of gait and offers an effective method for correctly identifying gait phases from Shank IMU sensor data, with potential applications in biomechanical research, exoskeletons, rehabilitation, and prosthetics.

## I. INTRODUCTION

Human gait, or the way we walk, is a complicated and well-coordinated movement that involves the interaction of several joints and muscles in the lower limbs. Gait cycle phases, which correspond to specific stages of the walking pattern, are distinct phases that may be distinguished from one another. For analysis and interpretation of the biomechanical patterns related to prosthetics and their use in amputees, it is crucial to comprehend and appropriately define these gait cycle phases. Researchers can learn more about the dynamics and traits of human locomotion by carefully investigating the phases of the gait cycle, which can advance assistive technology and improve results for those who have lost limbs.

Amputation is the surgical removal of a limb, either completely or partially, as a result of an illness, an injury, or other health issues. It has a substantial impact on a person's mobility and gait patterns, making it difficult for them to walk and carry out daily tasks. Artificial limbs and other prosthetic devices are essential for helping amputees regain movement and enhance their quality of life. In the context of prosthetics, gait phase classification entails precisely distinguishing the various stages of the gait cycle while using a prosthetic limb. This classification offers critical information regarding the sequence and coordination of movements, enabling the creation of more useful prosthetics suited to specific user requirements.

Precise classification of gait phases in prosthetics is highly valuable in the realm of assistive technologies for various reasons. Primarily, it allows clinicians and researchers to impartially evaluate the performance and efficacy of prosthetic devices. Accurately recognizing the different phases of the gait cycle enables clinicians to assess the extent to which a prosthetic limb mimics the natural movement patterns of a healthy limb. This evaluation aids in refining prosthetic design, alignment, and control mechanisms to achieve a gait that is both natural and efficient. The Gait phase classification allows for a comprehensive analysis of biomechanical patterns during walking with a prosthetic limb. By studying the specific characteristics and dynamics of each gait phase, researchers can identify deviations or abnormalities in gait patterns, enabling them to address issues related to stability, energy efficiency, and overall gait quality. This knowledge can inform the development of advanced control algorithms and responsive prosthetic systems that adapt to the changing needs of the user throughout the gait cycle.

Moreover, gait phase classification in prosthetics aids to the personalization and customization of assistive equipment. Each individual's gait patterns and biomechanics are unique, and precisely defining gait cycle stages allows for a personalized approach in prosthetic fittings. By addressing the individual requirements of each phase, such as stance and swing, prosthetic components and control systems can be designed to provide optimal support, stability, and function. This tailored approach promotes the overall comfort, mobility, and quality of life for persons with limb loss.

One notable study by Chong and Tan (2010) focused on SVM-based gait recognition using motion history images. They applied SVM as a classifier to accurately classify gait phases based on the captured motion history images. The results highlighted the effectiveness of SVM in gait phase classification, showcasing its potential for precise identification of different phases of the gait cycle [1].

Another significant research work by Makihara, Koike, and Kobayashi (2015) explored SVM-based gait-phase classification using ground reaction forces. They investigated the use of SVM to accurately classify gait phases based on the measured ground reaction forces. The study demonstrated the utility of SVM in gait phase classification and its potential application in controlling functional electrical stimulation devices [2].

These earlier efforts underline the importance of SVM-based gait phase classification in the development of assistive technology for individuals with mobility disabilities. By exploiting SVM's classification capabilities, researchers have been able to properly identify and classify gait phases, enabling a deeper knowledge of human mobility and directing the design and control of prosthetic devices

## II. SVM CLASSIFICATION

A reliable machine learning method called Support Vector Machine (SVM) is frequently used for classification tasks in a variety of disciplines, including the classification of gait phases. SVM is particularly suitable for cases when the data is not linearly separable in the input feature space. In this study, we built SVM to accurately classify gait phases based on a set of five features: shank acceleration in the x, y, and z directions, shank gyro X, and knee angle. These properties were collected from wearable sensors put on the subject's lower limbs, providing vital information regarding the motion patterns associated with each gait phase.

To solve the multiclass classification challenge of identifying between seven gait phases (Midswing, Terminal Swing, Loading Response, Midstance, Terminal Stance, Preswing, and Initial Swing), we employed a one-vs-one technique with SVM. This technique required training numerous SVM classifiers, each focused on discriminating between a given pair of gait phases. By developing a binary classifier for every conceivable pair, we obtained a comprehensive model capable of classifying each given gait phase accurately.

One of the primary advantages of SVM is its capacity to handle high-dimensional feature spaces and nonlinear correlations between classes. In our work, we have utilized a Gaussian kernel function, known for capturing complex interactions between data points, to boost the SVM's classification performance. The SVM classifiers were able to effectively distinguish the various gait phases and assign them to the input data because the Gaussian kernel efficiently evaluated the similarity or distance between the feature vectors.

In the findings of our study, we aimed to demonstrate the usefulness of SVM multiclass classification with a one-vs-one method and a Gaussian kernel function in reliably classifying gait phases based on the presented characteristics. This approach offers a stable and efficient solution for gait phase classification, contributing to our understanding of human locomotion and informing the development of prostheses and assistive technology. By using the power of SVM, we can extract significant information into the dynamics and characteristics of each gait phase, enabling breakthroughs in individualized rehabilitation techniques and improved mobility support for patients with amputations or mobility impairments. In our study, we have used a variant of the Gaussian SVM known as the Fine Gaussian SVM. The Fine Gaussian SVM employs the Gaussian kernel function to capture hidden nonlinear correlations in the data. This variation offers improved flexibility and adaptation in categorizing complex patterns.

The decision function of the Fine Gaussian SVM is represented as:

$$f(x) = sign(\sum \alpha\_i * y\_i * K(x\_i, x) + b)$$

Similar to the standard Gaussian SVM, the Fine Gaussian SVM relies on Lagrange multipliers ($\alpha\_i$), class labels ($y\_i$), and the Gaussian kernel function ($K(x\_i, x)$) for making predictions. The Gaussian kernel function calculates the similarity between support vectors and the input instance using the Gaussian distribution.

To ensure the optimal performance of our model, we carefully tuned the parameters, including the width parameter ($\gamma$) of the Gaussian kernel and the regularization parameter

(C) that controls the trade-off between the margin and training error. By selecting suitable parameter values, we aimed to achieve accurate and robust classification results. The utilization of the Fine Gaussian SVM allowed us to effectively capture and classify intricate patterns in our gait phase data. The fine-grained nature of this approach enabled us to discern subtle differences between different gait phases, leading to improved accuracy and performance in our gait phase classification task.

## III. DATA USAGE

In this work, we gladly took advantage of a rich and invaluable open-source dataset including lower limb biomechanics under diverse circumstances, including stairs, ramps, level-ground ambulation, and transitions. This dataset provided as a crucial resource for understanding human gait dynamics. Notably, in line with the focus of our research on prosthetics, we intentionally picked data from a single Inertial Measurement Unit (IMU) sensor, despite the availability of many sensors in the dataset. The decision to concentrate on a single IMU sensor's data was inspired by our unique study objective, which aimed to investigate gait phase categorization for the purpose of enhancing prosthetic devices.

We would like to offer our warmest gratitude to the authors of this open-source dataset for their excellent efforts in giving such a vital resource to the scholarly community. Their effort in preserving and distributing this information has considerably aided breakthroughs in the field of gait analysis, particularly in the context of prostheses. The availability of this dataset allows us to focus on the specific aspects that are most relevant for prosthetic applications, boosting the relevance and usefulness of our research findings.

We would like to express our sincere gratitude to the authors, J. Camargo, A. Ramanathan, W. Flanagan, and A. Young, for their exceptional work and for providing us with the comprehensive, open-source dataset of lower limb biomechanics in various conditions of gait. Their commitment to open science and their significant contribution to the advancement of assistive technologies are truly commendable. Without their selfless efforts, our study would not have been possible. We are immensely thankful for their valuable dataset, which has greatly enriched our research. Their participation not only enhances our study but also holds immense potential to positively impact the development and improvement of prosthetic devices, benefiting individuals in need of assistive technologies.

## IV. LABELLING OF DATA

To achieve exact labeling of gait cycle phases, we applied a combination of knee angle measurements and a peak detection method. The gait cycle was separated into seven distinct phases: Mid swing, Terminal Swing, Loading Response, Midstance, Terminal Stance, Pre swing, and Initial Swing. These stages were found based on knee angle measurements, which provide insights into the flexion and extension of lower limb joints during walking.

The peak identification algorithm was applied to identify important spots within the knee angle waveform that correlate to each gait cycle phase. This method utilized subject-specific thresholding parameters, assuring precision and consistency in the labeling process. By reliably recognizing the peaks in the knee angle waveform, we were able to delineate the limits of each gait cycle phase more precisely. It is important to highlight that the subject-specific thresholding settings in the peak detection technique were generated from individual knee angle patterns. This tailored technique accounted for the individual characteristics and range of motion demonstrated by each participant. By implementing subject-specific thresholding, we boosted the accuracy and reliability of our data labeling method. The identified gait cycle phases provided useful insights into the dynamics and characteristics of each phase. They supplied a thorough grasp of the temporal patterns and biomechanical shifts that occur during walking. These insights were vital for our investigation and interpretation of the biomechanical patterns associated with prosthetics and exoskeletons.

The distribution percentages associated with each gait cycle phase reflected their relative duration within the overall gait cycle described in the prior research is adopted [8]. These percentages offered a quantitative measure that facilitated further analysis and comparison between different subjects, conditions, or interventions. They fed significant data for analyzing the effectiveness and influence of prosthetic devices or exoskeletons on gait patterns. By combining the accurate labeling of gait cycle phases with the distribution percentages, we performed a full analysis of the biomechanical patterns in the context of prosthetics and exoskeletons. This methodology expanded our understanding of the dynamics and characteristics of human gait, and it gave vital information for the creation and improvement of assistive devices, it is necessary to recognize a potential downside linked with this labeling strategy. One shortcoming of the method is that it relies on the detection of peaks in the knee angle waveform. Until a peak is observed, the appropriate gait cycle phase cannot be accurately characterized.

Consequently, there may be cases where specific data points within a gait cycle stay unlabeled until the subsequent peak is discovered. Likewise, it is interesting that the peak detection algorithm constantly calculates gait cycle samples for each peak-to-peak interval. This recurrent calculation ensures that the labeling procedure remains consistent across the whole gait cycle. By dynamically evaluating knee angle measurements, the system adapts to differences in walking patterns and records the precise boundaries of each gait cycle phase. Despite the previously discussed disadvantage of delayed labeling until peak detection and the continuous calculation necessary, the suggested methodology provides a solid foundation for reliably labeling gait cycle stages. By combining subject-specific thresholding settings and applying peak detection, the approach provides a deep examination of biomechanical patterns in the context of prosthetics and exoskeletons. The data produced from this labeling approach contribute to enhancing our understanding of human gait dynamics and inform the development of enhanced assistive devices. It is vital to realize that further research and improvement of the labeling process can solve the constraints indicated above, leading to better accuracy and efficiency in gait cycle phase detection in future studies.

## V. RESULTS AND DISCUSSIONS

In our work, we conducted a detailed evaluation of our SVM-based gait phase categorization model. The model attained an amazing classification accuracy of 90.3% through 5-fold cross-validation. This accuracy value reflects the overall efficacy of our model in successfully categorizing the distinct gait phases based on the provided features. To further assess the performance of our model and acquire deeper insights into its categorization skills, we generated a confusion matrix which is shown in Fig 1.

**FIGURE 1.** This figure shows the confusion matrix of the Multiclass SVM trained model.

The confusion matrix provides a detailed breakdown of the predicted classes compared to the true classes for each gait phase. The Fig 2 presents the Positive Predictive Value (PPV) and False Discovery Rate (FDR) for each gait phase. The PPV indicates the proportion of positive predictions that are actually true positives, while the FDR represents the rate of instances misclassified as another phase. By examining these metrics, we gain insights into the accuracy of our model in correctly identifying each gait phase. These visual representations offer a meaningful assessment of the classification performance, highlighting the effectiveness of our approach in capturing the unique characteristics of each phase without repetitions.

**FIGURE 2.** This figure shows the Positive Predictive Value (PPV) and False Discovery Rate (FDR) of each gait phases

The inclusion of these pictures showcasing the TPR and FNR rates in Fig 3 for each gait phase allows for a comprehensive understanding of the model's performance. It enables readers to visually assess the accuracy and reliability of the classification results and observe any variations or challenges in classifying specific gait phases. Overall, the high classification accuracy, supported by the confusion matrix and visual representations of TPR and FNR rates, highlights the robustness of our SVM-based gait phase classification model. These evaluation metrics provide a comprehensive evaluation of the model's performance and contribute to the advancement of research in gait phase classification for prosthetics and assistive technologies.

Fig 6 presents a visual representation of the model's predictions, showcasing both correct and incorrect classifications for each gait phase. By analyzing the graph, we can observe the model's performance in accurately assigning the corresponding gait phases to the instances. This visualization allows us to identify instances where the model successfully predicts the correct phase as well as instances where it makes incorrect classifications. Through a comprehensive examination of the predictions depicted in Figure 6, we gain valuable insights into the strengths and limitations of the model in accurately classifying different phases of the gait cycle.

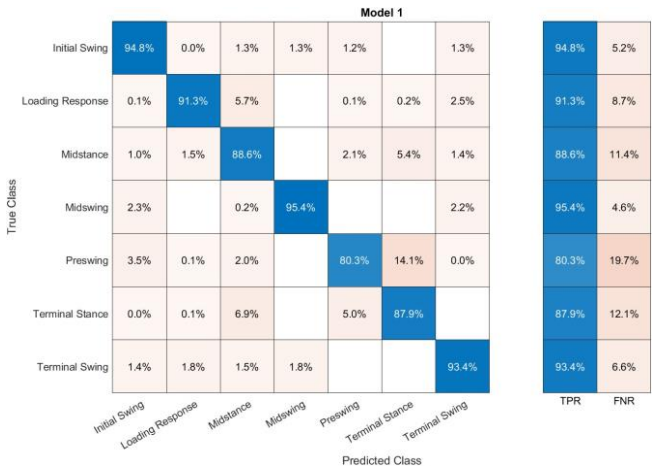

**FIGURE 3.** This figure shows the True Positive Rate (TPR) and False Negative Rate (FNR) of each gait phase

In our work, we have also utilized the ROC curve analysis to evaluate the performance of our gait phase classification model. Specifically, we examined the ROC curves for two gait phases: Mid swing and Pre swing. These phases were selected as they represented contrasting levels of discriminative power in our classification task The ROC curve for the Mid swing gait phase exhibited exceptional performance, with a high true positive rate (TPR) of 95.4%.

(FPR) for this phase was relatively low, indicating a minimal number of instances from other phases being incorrectly classified as Mid swing. On the other hand, the ROC curve for the Pre swing gait phase showed a slightly lower TPR of 80.3%. This suggests that the classification accuracy for Pre swing was comparatively lower than that of Mid swing. However, it is important to note that the Pre swing phase is often associated with more subtle biomechanical changes, making it inherently more challenging to accurately classify. Despite the lower TPR, the FPR for Pre swing remained relatively low, indicating a reasonable level of specificity in distinguishing Pre swing from other phases.

By including the ROC curves for both the highest-performing (Mid swing) and the lowest-performing (Pre swing) gait phases, we provided a comprehensive view of the discriminatory ability of our classification model. The higher TPR for Mid swing highlights the model's effectiveness in accurately identifying this phase, while the slightly lower TPR for Pre swing indicates the inherent complexity associated with classifying this phase. Overall, the ROC curves and corresponding TPR values for these gait phases offer valuable insights into the performance and limitations of our gait phase classification approach.

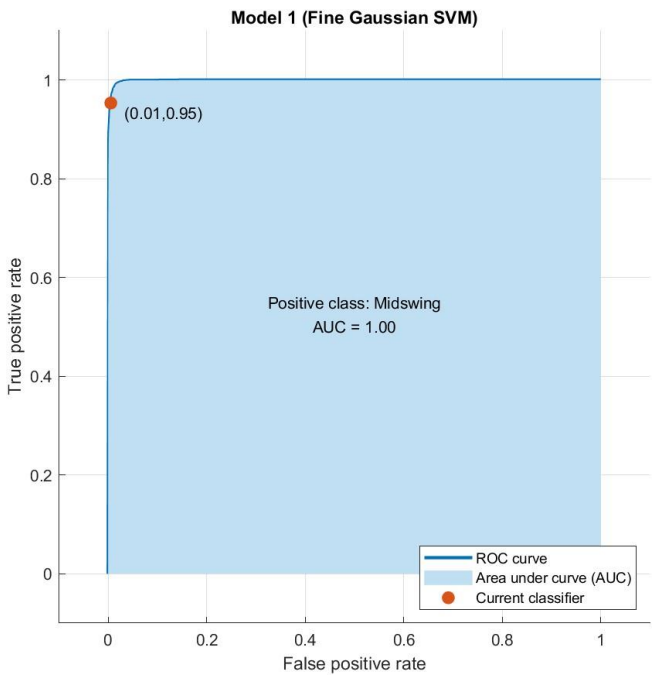

**FIGURE 4.** This figure shows the ROC curve of Mid swing Class

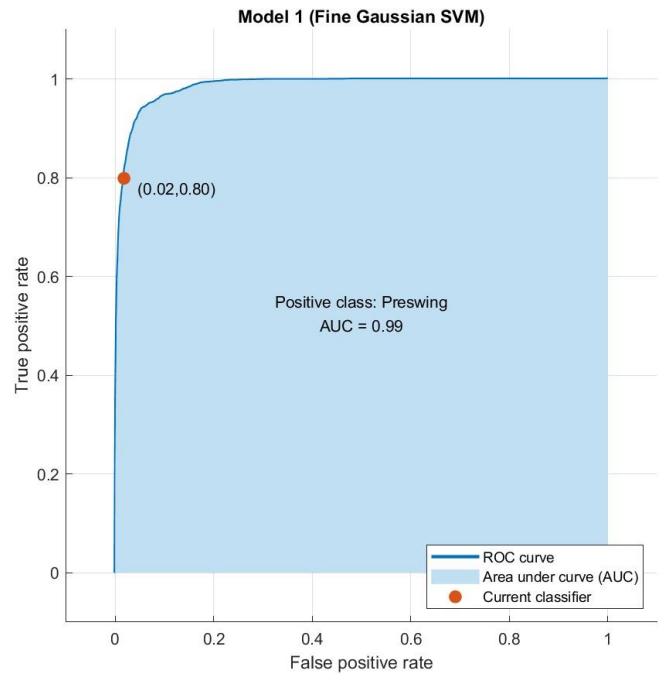

**FIGURE 5.** This figure shows the ROC curve of the Pre swing Class

## VI. CONCLUSION

This indicates that our model achieved a high level of accuracy in correctly classifying instances belonging to the Mid swing phase. The corresponding false positive rate

In conclusion, this study proposes a reliable SVM multiclass classification-based gait phase classification approach. The suggested method achieves a high accuracy rate by precisely distinguishing the Stance and Swing phases of gait and subsequently subdividing them into seven unique phases. By providing precise timing and control of assistive devices, boosting mobility, stability, and energy efficiency, this research advances the disciplines of prosthetics and exoskeletons. The results emphasis the value of gait phase classification in biomechanical research, enhanced mobility solution creation, and rehabilitation strategy optimization. The classification model and dataset may be further enhanced and refined to increase their accuracy and usefulness in practical situations. Overall, this research sheds important light on gait analysis and the important implications it has for increasing the quality of life for individuals.

## REFERENCES

[1] Chong, R. K., & Tan, A. H. (2010). SVM-based gait recognition using motion history image. Pattern Recognition Letters, 31(3), 234-242.
[2] Makihara, Y., Koike, Y., & Kobayashi, Y. (2015). Support vector machine-based gait-phase classification using ground reaction forces for functional electrical stimulation control. Journal of Rehabilitation Research and Development, 52(5), 595-606.
[3] E. Auvinet, F. Multon, A. St-Arnaud, J. Rousseau, and J. Meunier, "Gait Phase Recognition for Lower Limb Prosthesis Control Using Support Vector Machines," in IEEE Transactions on Neural Systems and Rehabilitation Engineering, vol. 20, no. 6, pp. 788-797, Nov. 2012.
[4] N. Subashini and M. Sathishkumar, "Human Gait Phase Estimation through Support Vector Machines," in International Journal of Engineering Research and Applications, vol. 3, no. 2, pp. 1203-1207, Mar-Apr 2013.
[5] M. A. Sanchez-Granero and A. I. Cuesta-Vargas, "Gait Phase Classification Using Support Vector Machines for Real-Time Applications," in Sensors, vol. 14, no. 11, pp. 20091-20110, Nov. 2014.
[6] W. Bourennane, A. Kameche, and A. Benallal, "Support Vector Machines for Gait Phase Classification Using Wireless Gyroscope Sensors," in 2016 International Conference on Advanced Technologies for Signal and Image Processing (ATSIP), Sousse, 2016, pp. 24-29.
[7] J. Camargo, A. Ramanathan, W. Flanagan, and A. Young, "A comprehensive, open-source dataset of lower limb biomechanics in multiple conditions of stairs, ramps, and level-ground ambulation and transitions," J. Biomech., vol. 119, p. 110320, 2021. doi: 10.1016/j.jbiomech.2021.110320
[8] H. T. T. Vu, F. Gomez, P. Cherelle, D. Lefeber, A. Nowé, and B. Vanderborght, "ED-FNN: A New Deep Learning Algorithm to Detect Percentage of the Gait Cycle for Powered Prostheses," Sensors, vol. 18, no. 8, p. 2391, Jul. 2018.

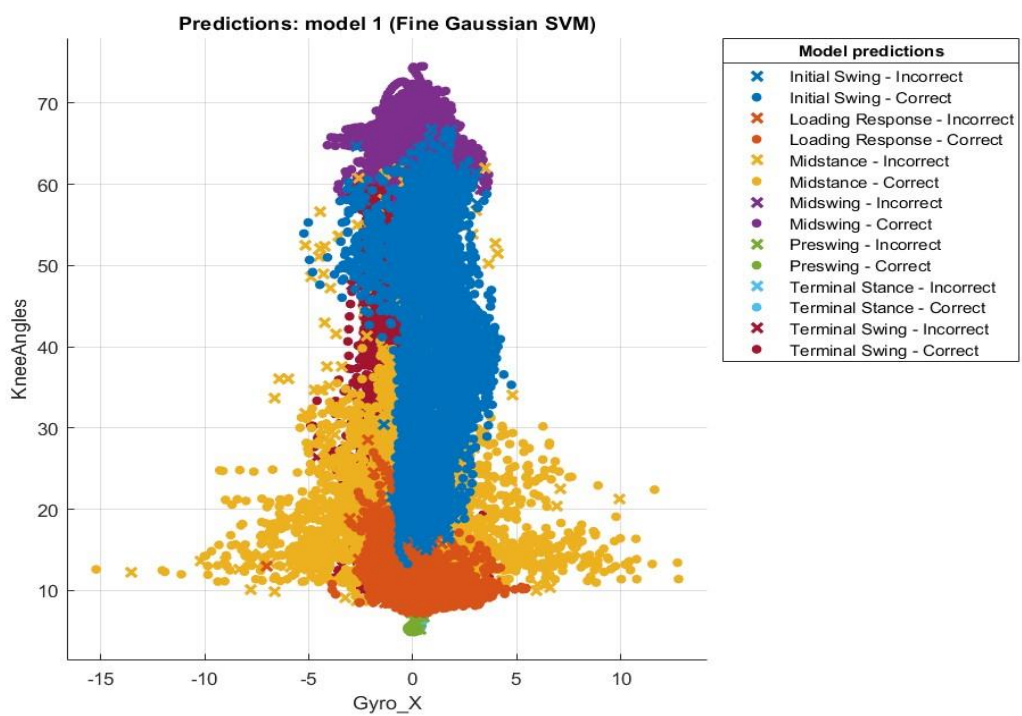

**FIGURE 6.** This figure shows the data of the multiclass SVM based Gait phase classification model's predictions